\documentclass[10pt,journal,compsoc]{IEEEtran}
\pdfoutput=1

\usepackage{times}
\usepackage{epsfig}
\usepackage{graphicx}
\usepackage{amsmath}
\usepackage{amssymb}
\usepackage{tabularx,colortbl,multirow,float}
\usepackage{comment}
\usepackage{enumitem}
\setlist{nosep}

\DeclareMathOperator*{\argmax}{arg\,max}

\ifCLASSOPTIONcompsoc
  
  \usepackage[nocompress]{cite}
\else 
  \usepackage{cite}
\fi

\ifCLASSINFOpdf
  
\else
  
\fi

\begin{document}

\onecolumn 

\begin{description}[leftmargin=2cm,style=multiline]

\item[\textbf{Citation}]{M. Prabhushankar, and G. AlRegib, "Contrastive Reasoning in Neural Networks," submitted to \emph{IEEE Transactions on pattern analysis and machine intelligence}, Jan. 2021.}

\item[\textbf{Copyright}]{\textcopyright 2020 IEEE. Personal use of this material is permitted. Permission from IEEE must be obtained for all other uses, in any current or future media, including reprinting/republishing this material for advertising or promotional purposes,
creating new collective works, for resale or redistribution to servers or lists, or reuse of any copyrighted component
of this work in other works.}

\item[\textbf{Contact}]{mohit.p@gatech.edu OR alregib@gatech.edu \\ https://ghassanalregib.info/}

\end{description}

\thispagestyle{empty}
\newpage
\clearpage
\setcounter{page}{1}

\twocolumn

\title{Contrastive Reasoning in Neural Networks
}

\author{Mohit Prabhushankar,~\IEEEmembership{Student Member,~IEEE,} 
        and~Ghassan~AlRegib,~\IEEEmembership{Senior Member,~IEEE}
\IEEEcompsocitemizethanks{\IEEEcompsocthanksitem Mohit Prabhushankar is with the School
of Electrical and Computer Engineering, Georgia Institute of Technology, Atlanta,
GA, 30332.\protect\\
E-mail: mohit.p@gatech.edu
\IEEEcompsocthanksitem Ghassan AlRegib is with the School
of Electrical and Computer Engineering, Georgia Institute of Technology, Atlanta,
GA, 30332.\protect\\
E-mail: alregib@gatech.edu}
\thanks{Manuscript submitted on Jan 09, 2021.}}

\markboth{Submitted to IEEE Transactions on Pattern Analysis and Machine Intelligence}%
{Prabhushankar \MakeLowercase{\textit{et al.}}}

\IEEEtitleabstractindextext{%
\begin{abstract}
Neural networks represent data as projections on trained weights in a high dimensional manifold. The trained weights act as a knowledge base consisting of causal class dependencies. Inference built on features that identify these dependencies is termed as feed-forward inference. Such inference mechanisms are justified based on classical cause-to-effect inductive reasoning models. Inductive reasoning based feed-forward inference is widely used due to its mathematical simplicity and operational ease. Nevertheless, feed-forward models do not generalize well to untrained situations. To alleviate this generalization challenge, we propose using an effect-to-cause inference model that reasons abductively. Here, the features represent the change from existing weight dependencies given a certain effect. We term this change as contrast and the ensuing reasoning mechanism as contrastive reasoning. In this paper, we formalize the structure of contrastive reasoning and propose a methodology to extract a neural network's notion of contrast. We demonstrate the value of contrastive reasoning in two stages of a neural network's reasoning pipeline : in inferring and visually explaining decisions for the application of object recognition. We illustrate the value of contrastively recognizing images under distortions by reporting an improvement of $3.47\%$, $2.56\%$, and $5.48\%$ in average accuracy under the proposed contrastive framework on CIFAR-10C, noisy STL-10, and VisDA datasets respectively. 
\end{abstract}

\begin{IEEEkeywords}
Abductive Reasoning, Contrastive Explanations, Contrastive Inference, Knowledge representation, Robustness 
\end{IEEEkeywords}}

\maketitle

\IEEEdisplaynontitleabstractindextext
\IEEEpeerreviewmaketitle

\IEEEraisesectionheading{\section{Introduction}\label{sec:introduction}}

\IEEEPARstart{S}{upervised} learning entails learning by association. In the field of psychology, associative learning involves the formation of previously unknown associations between stimuli and responses~\cite{byrne2014learning}. In supervised machine learning, stimuli refers to the data provided to the algorithm and response is the label required of the algorithm. During training, the algorithm is conditioned to associate between the data and its label. Hence, the goal of training an algorithm is to learn the patterns in data that either cause or correlate to the given label. During testing, the algorithm searches for the learned pattern to predict the label. This act of predicting the label is termed as inference. The means of arriving at the inference is reasoning. An example of supervised machine learning is object recognition where a label is inferred based on the learned patterns in a given image.

Recent advances in machine learning has allowed state-of-the-art performances in recognition tasks~\cite{krizhevsky2012imagenet}\cite{he2016deep}. Specifically, recognition algorithms surpassed top-5 human error rate of $5.1\%$ on ImageNet~\cite{russakovsky2015imagenet}. Error rate measures the inaccurate predictions made by humans or the network. The reduction in error rate can be traced to advancements in the learning process~\cite{kingma2014adam}, regularization~\cite{srivastava2014dropout}, and hardware utilization~\cite{li2017analysis} among others. However, all these advancements rely on the traditional feed-forward-based associative reasoning and inference scheme. The author in~\cite{olshausen201427} examines the conventional feed-forward model of inference in both human vision and machine vision, where a set of neurons extract object selective features in a hierarchical fashion. Such a feed-forward representation leads to a task-specific, trained-data specific inference mechanism. For instance, a change in data domain during testing requires retraining to obtain a new representation conducive to test domain. This is because the feed-forward model follows an inductive reasoning approach to inference. Inductive reasoning is a branch of causal inference in perception~\cite{shams2010causal}. Inference based on induction provides a conclusion with uncertainty which allows for speculation regarding cause. 

\begin{figure*}[!htb]
\begin{center}
\minipage{1\textwidth}%
  \includegraphics[width=\linewidth]{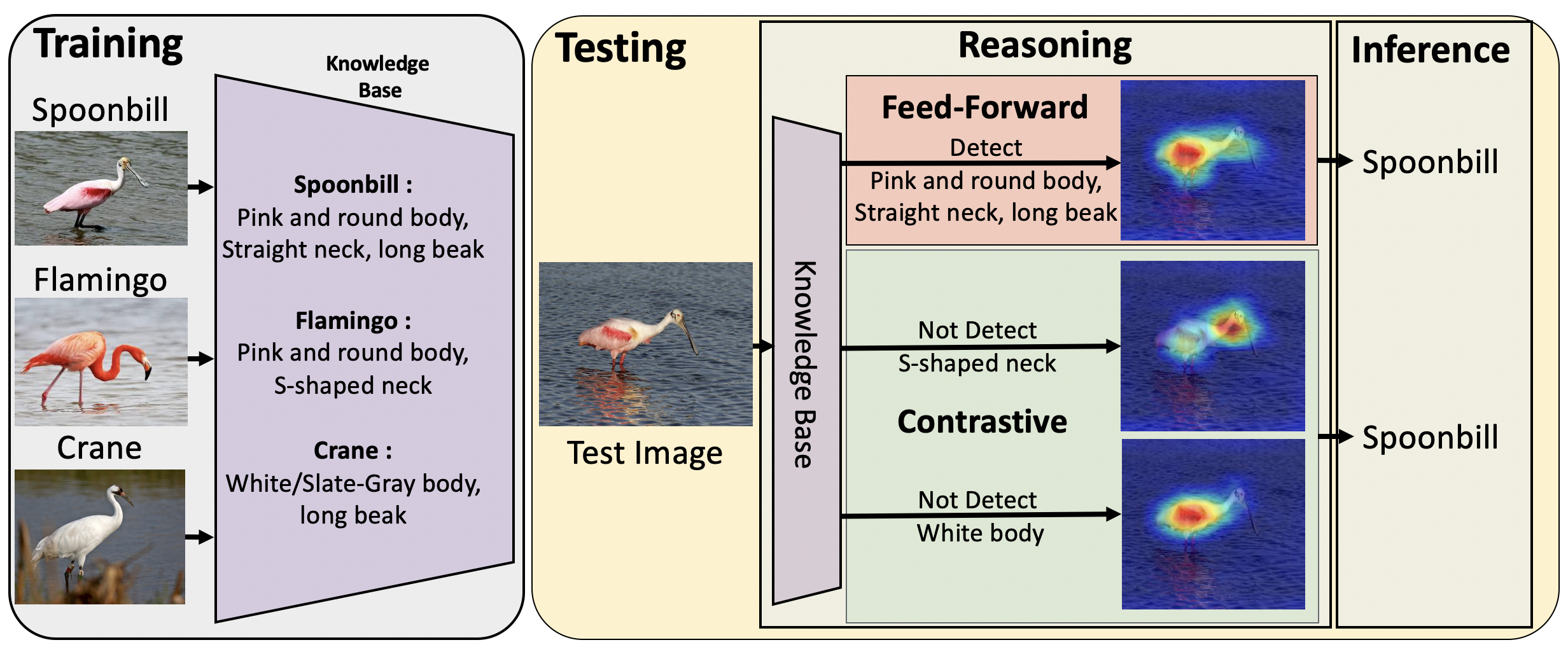}
\endminipage
\caption{Discriminative features are identified in Feed-Forward reasoning frameworks. Contrasts between observed facts and a known knowledge base is used as reasons in Contrastive Inference.}\label{fig:Concept}
\end{center}
\end{figure*}

Contrary to an inductive framework, an abductive framework creates a hypothesis and tests its validity without considering the cause. Abductive reasoning allows humans to better generalize to new situations. Extensive work has been conducted to understand the development of human brain and visual perception based on abductive reasoning~\cite{national2018people}. This form of reasoning was introduced by the philosopher Charles Sanders Peirce~\cite{peirce1931collected}, who saw abduction as a reasoning process from effect to cause~\cite{paul1993approaches}. In contrast, induction conjectures general laws from particular instances. Peirce further connected induction and abduction by stating that \emph{Induction is an argument which sets out from a hypothesis, resulting from previous Abduction}. In this paper, we follow this principle to modify inductively trained feed-forward neural networks to reason and infer abductively for the task of object recognition. In other words, we use an inductively trained neural network to extract multiple hypotheses regarding the input's class and use them as features for inference. Specifically, we extract hypotheses based on the network's assertion of contrast between classes. We call this inference mechanism as \emph{contrastive inference} and its corresponding reasoning as \emph{contrastive reasoning}.

To explain this concept of contrast, let us take the toy example of two subjects distinguishing between three classes of birds - spoonbill, flamingo and crane. This is illustrated in Fig.~\ref{fig:Concept}. All three are shallow-water birds with fine-grained physical differences. Spoonbills and flamingos have round and pink bodies. However, they differ primarily in the shape of their necks. A flamingo has an S-shaped neck while a spoonbill does not. Spoonbills and Cranes have long necks but have different colored bodies. Consider that images of all three birds have been shown to two subjects beforehand, thereby training them to distinguish between the three birds. Given a new test image of a spoonbill, subject A correctly recognizes the image as a spoonbill based on the shape and color of the bird's body and its beak. Subject B also recognizes the image as a spoonbill but does so by noticing that the neck is \emph{not} S-shaped and the body is \emph{not} white. While both the subjects infer correctly, the reasoning mechanism is complementary. Subject B infers the \emph{contrast} between a known fact - the S-shape of the neck, white body- and its lack thereof in the test image to make the right classification. Hence subject B constructed multiple hypotheses and contrastively reasoned why it cannot be either a flamingo or a crane. Inference that is built on these contrastive reasons or features is termed as \emph{contrastive inference}. On the contrary, subject A reasons inductively to infer the correct class based on identifying the patterns in the image corresponding to spoonbill. This is feed-forward inference that is currently being used by neural networks. 

\noindent \textbf{Paper structure and novelty: }In this paper, we rethink existing inductive reasoning based feed-forward networks and provide an alternative reasoning and inference scheme based on abduction. We first describe abductive reasoning and its usage in artificial intelligence and machine learning in Section~\ref{Sec:LitReview}. We provide a structure to the possible abductive reasons and introduce contrastive reasons as a subset of abductive reasons. We further show that current formulations of abductive reasoning does not allow for large scale learning based solutions. The novelty of our work is in formulating abductive reasons as contrasts that can be extracted from trained deep networks. This is motivated and described in Section~\ref{sec:Abduction}. We extract and visualize these contrastive reasons in Section~\ref{Sec:Contrast} as contrastive visual explanations. We then qualitatively examine the results and gain insights provided by such contrastive explanations. In Section~\ref{Sec:Inference}, we describe the inference mechanism that uses the contrastive reasons to make decisions. We finally enhance the robustness of existing models to noise in Section~\ref{sec:Results} and conclude in Section~\ref{Sec:Discussion}. 
\section{Related Works}
\label{Sec:LitReview}
In Section~\ref{sec:introduction}, we introduced abductive reasoning as an alternative reasoning scheme to existing induction that better generalizes to new situations. Abductive reasoning requires multiple hypotheses to infer decisions. Hypotheses are defined as answers to logical questions that aid inference. In this section, we first describe the possible logical questions - causal, counterfactual, and contrastive. We then motivate our choice of contrastive questions to obtain abductive hypotheses.
\subsection{Abductive Reasoning}
\noindent \textbf{Abductive Reasoning: }Consider a classification network where given a datapoint $x$ the network predicts the label of $x$ as $P$. The factors that cause $P$ are termed causal reasons~\cite{pearl2009causal, halpern2005causes}. An explanation for the generation of label $P$ can be defined as the answer to the question \emph{'Why P?'}. For instance, for a classification algorithm, Grad-CAM~\cite{selvaraju2017grad} provides visual explanations for \emph{'Why P?'}. However, these explanations are based on observational causality~\cite{lopez2017discovering}. Observed causality can hence be differentiated from interventionist causality where the explanations for a predicted label $P$ changes in response to active interventions. Active interventions can further be divided into two scenarios based on the location of interventions - either in the data itself or in the predictions~\cite{mcgill1993contrastive}. Intervening within data itself provides answers to \emph{'What if?'} questions. These answers are counterfactual explanations. However, all possible interventions in data can be long, complex and impractical~\cite{lopez2017discovering}. Even with interventions, it is challenging to estimate if the resulting effect is a consequence of the intervention or due to other uncontrolled interventions~\cite{bottou2013counterfactual}. The second type of interventions can occur in the network predictions. These are questions of the type \emph{'Why P, rather than Q?'} questions. The answers to these questions form contrastive reasons. The answers to observed causal, counterfactual, and contrastive questions constitute abductive reasons. Before describing the current formulations of abductive reasoning and frameworks, we relate the considered observed causal and contrastive reasons to Fig.~\ref{fig:Concept}.

\noindent \textbf{Abductive and Contrastive Reasons: }In this paper, we specifically choose contrastive questions to obtain abductive hypotheses and extract answers from neural networks by intervening in the predictions made. This is because pre-trained neural networks already have implicit contrastive knowledge. We first motivate this claim. In Fig.~\ref{fig:Concept}, consider that the knowledge base is a trained neural network $f()$. During testing, it is given the image of a spoonbill $x$, with the task of predicting the label $P$ of $x$. Consider that $f()$ correctly predicts the label $P$ as spoonbill. The feed-forward reason for $P$ is the detection of pink and round body, and straight neck. These are reasons for \emph{`Why Spoonbill?'} and represent observed causal reasons. The contrastive reasons answer the question \emph{`Why Spoonbill, rather than flamingo?'} and \emph{`Why spoonbill, rather than crane?'}. Here, $Q$ is either flamingo or crane. The contrastive reasons in Fig.~\ref{fig:Concept} visualize the knowledge base's notion of the difference between a spoonbill and a flamingo and between a spoonbill and a crane. While $P$ is constrained by the prediction from $f()$, $Q$ is user defined and can be used to ask the network for other explanations including \emph{`Why spoonbill, rather than band-aid?'} or \emph{`Why spoonbill, rather than pig?'} as long as the network $f()$ has learned these classes. $Q$ can be any of the learned classes in the network. If $f()$ is trained to classify between $N$ classes, $Q$ can take the values of $[1, N]$. Hence, for a recognition network with some prediction $P$ and trained to discriminate between $N$ classes, there are potentially $N$ possible contrastive reasons. The network acts as a discriminatory knowledge base that has implicitly learned the contrastive information between $P$ and $Q$. Therefore, we choose the contrastive model of abductive reasoning in our framework. 

\noindent\textbf{Reasoning versus Explanations: }Reasoning is the process by which a network $f()$ infers a decision $P$. The authors in~\cite{goguen1983reasoning} consider reasoning to be a mental process which can only be surmised based on how it manifests - in terms of explanations and inference. Explanations are comprehensible justifications of reasoning. In other words, explanations are justifications that a network $f()$ provides after a decision $P$ is made. Inference is the outcome of reasoning, i.e. the decision $P$ itself. Therefore, for all trained networks, we can ask and extract observed causal~\cite{selvaraju2017grad}, counterfactual~\cite{goyal2019counterfactual}, and contrastive explanations~\cite{prabhushankar2020contrastive}. Similarly, inference can occur based on causal, counterfactual, or contrastive features. They can then be termed as causal, counterfactual, or contrastive inference. Two of these explanatory schemes and inference mechanisms are shown in Fig.~\ref{fig:Concept}. Technically, the explanations come after decisions but we use explanations as a visual substitute for reasoning in Fig.~\ref{fig:Concept}. 
\subsection{Abductive Inference and Learning}
\noindent \textbf{Abductive Inference: }Abductive inference is generally described as Inference to the Best Explanation (IBE)~\cite{harman1965inference}. While inductive inference associates learned patterns and extrapolates them to unseen test cases, IBE postulates all possible explanations for the occurrence of the test case and picks the class with the \emph{best} explanation. Note that the novelty of existing works is derived from finding a metric that describes what explanation is \emph{best}. Traditionally, the simplicity of an explanation was seen as a stand in for the best explanation. Recent works for IBE in machine learning are postulated under the abductive framework. 

\noindent \textbf{Abductive Framework: }The combination of both abductive reasoning and inference together forms the abductive framework. Abductive reasoning is a common-sense based reasoning rather than mathematical reasoning. As such, its utility is in specific real world scenarios - police investigations~\cite{carson2009abduction}, medical diagnostics~\cite{magnani1992abductive}, automobile diagnostics~\cite{finin1989abductive}, archaeology~\cite{thagard1997abductive}, ontology~\cite{elsenbroich2006case}, and psychology~\cite{shank1998extraordinary} among others. In all these cases, the formulation of abduction is based on first-order logic. Modern deep learning algorithms have far exceeded the capabilities of first-order logic. In machine learning, an abductive framework is primarily seen as a logical process. The authors in~\cite{kakas2000abductive} describe the logical process of abduction and contrast it against an inductive logical process. The authors in~\cite{dai2019bridging} instantiate abductive reasoning as first-order logical program that runs in parallel to existing perception models to decipher mathematical equations in an image. The first-order program requires knowledge concepts that are input into the system apart from the data itself. They use a generic CNN as the perception model to recognize the symbols in the given image. An external abductive model is defined which consists of knowledge of the structure of equations, and the recursive bitwise operations. 

\subsection{Contrastive Inference and Learning}
In this paper, we divine the knowledge concepts intrinsically from the perception network. We do so by defining the knowledge concept as the difference between $P$ and $Q$ entities from the contrastive reasoning statement \emph{`Why P, rather than Q?'}. The name is derived from the psychological concept of contrastive inference. 

\noindent\textbf{Contrastive Inference: }In developmental psychology, contrastive inference is a mechanism that allows for the divination of entities that are not mentioned linguistically~\cite{kronmuller2014show}. For instance, the phrase '\emph{Hand me the large cup}' implies the existence of multiple cups of varying sizes without the need for mentioning their existence explicitly. Human-interpretable explanations have been shown to be contrastive - explanations are based on a contrast to a known fact. Contrastive inference provides a way of obtaining pragmatic explanations and hence is a form of abductive reasoning~\cite{kronmuller2014show,lipton2003inference}.

\noindent\textbf{Contrastive Learning: } The term contrastive learning denotes a framework wherein negative samples of training data that are semantically similar to positive samples are mined~\cite{arora2019theoretical}\cite{NIPS2013_5007}\cite{chen2020simple}. In~\cite{NIPS2013_5007}, contrastive learning is used to differentiate between specific topics within mixture models. The authors in~\cite{chen2020simple} propose a contrastive learning framework where multiple data augmentation strategies are used to train a network overhead with a contrastive loss. Note that the proposed method does not derive from the existing contrastive learning setup.

\noindent \textbf{Proposed Contrastive Inference: }We comment on the difference between the proposed contrastive inference and existing abductive frameworks. In any given scenario, an abductive inference framework makes one of $N$ possible decisions based on $N$ independent hypotheses that are extracted to verify the validity of $N$ separate decisions. The hypotheses are extracted based on a knowledge base. One of the hypotheses is then chosen and its corresponding decision made. Consider such a framework from~\cite{dai2019bridging} where an external model is a knowledge base. Consider such a model applied to large scale recognition datasets. In the example of Fig.~\ref{fig:Concept}, the knowledge base must be manually initialized with the neck and body category for existing reasoning mechanisms to function well. Consider the number of semantic categories requiring manual allocation if the differentiation is among $1000$ classes in ImageNet dataset. Initializing the knowledge base becomes both computationally tedious and defeats the purpose of learning from data. In this paper, we address this challenge by extracting contrast based abductive reasons directly from the network. We start by defining contrast both in the visual space and in the representation space spanned by the neural network.

\begin{figure*}
\begin{center}
\minipage{1\textwidth}%
  \includegraphics[width=\linewidth]{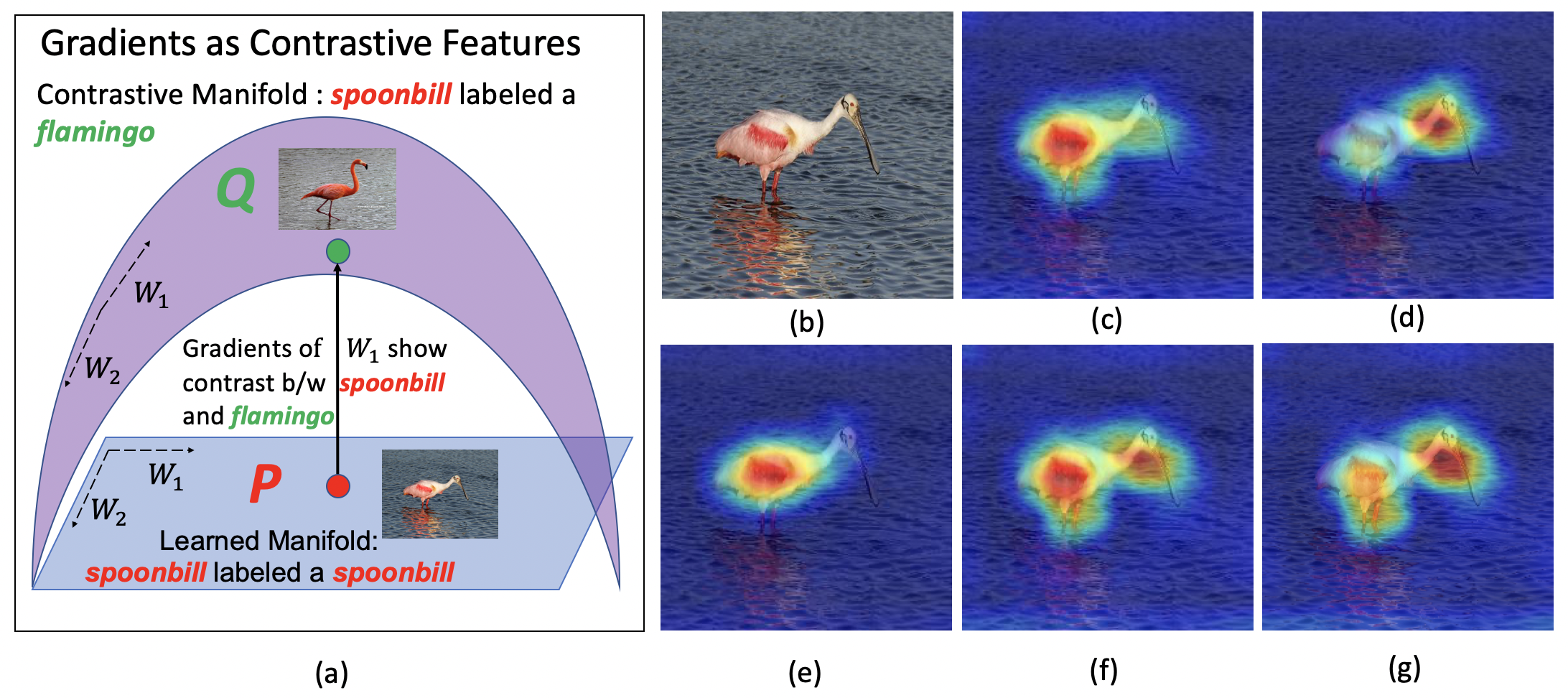}
\endminipage
\caption{Contrastive Feature Generation. (a) Learned manifold in blue is the original manifold that recognizes a spoonbill as a spoonbill. Contrastive manifold in purple is the manifold where a spoonbill is classified as a flamingo. Change between the two is termed contrast. (b) Spoonbill. (c) Grad-CAM~\cite{selvaraju2017grad} (d) Contrast between Spoonbill and Flamingo. (e) Contrast between Spoonbill and Crane. (f) Contrast between Spoonbill and Band-Aid. (g) Contrast between Spoonbill and a Pig.}\label{fig:Contrast_examples}
\end{center}
\end{figure*}

\section{Contrastive Features}
\label{sec:Abduction}
In visual space, we define contrast as the perceived difference between two known quantities. In this paper, we assume that the knowledge of the two quantities is provided by a trained neural network $f()$. The knowledge present in a feed-forward classification network is discriminatory and its reasoning process is inductive. In other words, given a neural network $f()$ that is trained to classify between $N$ classes, the network recognizes patterns to infer a class $P$ for any given image $x$. We first describe the feed-forward features used to make inductive decisions before providing a representation space definition for contrast.
\vspace{-3mm}
\subsection{Feed-Forward Features}\label{subsec:FF-Features}
Consider an $L$-layered classification network $f()$ trained on $x \in X$ for a certain task. $f()$ stores knowledge about the task on $X$ in the parameters of its layers - $W_l$ and $b_l$, $\forall l \in [1,L]$. Given an input image $x$, traditional feed-forward networks that reason inductively project the image on these parameters to make a decision $P$. The intermediate activations from a layer $l$ in $f()$ are termed $y_{feat}^l$. These activations are projections on the span of the weights $W_l, \forall l \in [1,L-1]$. $y_{feat}^l$ are the features used to make the decision $P$ and hence, activations $y_{feat}^l$ are the feed-forward features. In both the explanation and inference applications, we use $y_{feat}^l$ as feed-forward inductive features. $y^l_{feat}$ are used to obtain explanations for decisions using Grad-CAM~\cite{selvaraju2017grad} which is described in Section~\ref{Sec:Contrast}. If $y_{feat}^{L-1}$ are the feed-forward features at the $(L-1)^{th}$ layer, a task specific mechanism is used to infer the feed-forward prediction $P$ which is explored in Section~\ref{Sec:Inference}.
\vspace{-3mm}
\subsection{Contrastive Features}\label{subsec:Contrastive-Features}

Note that contrastive reasons are a subset of all possible abductive reasons. We adopt the definition of abduction from~\cite{velazquez2013epistemic} who define abduction as \emph{a process of belief change that is triggered by an observation and guided by the knowledge and belief that an agent has the ability to derive}. Hence, contrast is a measure of change created by an input image $x$ in the parameters of a trained network against the contrast class. In terms of neural network representation space, contrast between classes $P$ and $Q$ for an image $x$ is the difference between the manifolds that predict $x$ as class $P$ and $x$ as $Q$. The network parameters $W_l$ and $b_l$ span a manifold where the given image $x$ belongs to a class $i, i \in [1,N]$. A toy classification example is shown in Fig.~\ref{fig:Contrast_examples} where a learned manifold, $m_P$, is visualized in blue. On the learned manifold, a spoonbill is classified as a spoonbill. A hypothetical contrastive manifold, $m_Q$, is shown in purple that differs from the blue manifold in that it classifies a spoonbill as a flamingo. The difference between the two manifolds is contrast. Note that $m_Q$ is hypothetical and hence the difference between the two cannot be directly measured. In this paper, we measure the change required to obtain the contrastive manifold from the trained manifold. We use gradients to measure this change. Usage of gradients to characterize model change is not new. Neural networks whose objectives can be formulated as a differentiable loss function are trained using backpropagation~\cite{rumelhart1986learning}. The authors in~\cite{kwon2019distorted} used gradients with respect to weights to characterize distortions for sparse and variational autoencoders. Fisher Vectors use gradients to characterize the change that data creates within networks~\cite{jaakkola1999exploiting} which were extended to classify images~\cite{sanchez2013image}. In detection applications on anomalous~\cite{kwon2020backpropagated}, novel~\cite{kwon2020novelty}, and out-of-distribution~\cite{lee2020gradients} data, gradients are used to characterize model change.

We extract contrast for class $Q$ when an image $x$ is predicted as $P$ by backpropagating a loss between $P$ and $Q$. Hence, for a loss function $J()$, contrast is $\nabla_{W_l} J(W,x,P,Q)$, where $W$ are the network weights. Note that $J()$ is a measure of contrastivity between $P$ and $Q$. $Q$ can be any one of $[1,N]$ classes. Hence for $i \in [1,N]$, there are $N$ possible contrastive features given by $\nabla_{W_l} J(W,x,P,i)$ at any layer $l$ for $i \in [1,N]$. The feed-forward features $y^l_{feat}$ and the proposed contrastive features are analogous in the sense that they provide mechanisms to infer or justify decisions. In Sections~\ref{Sec:Contrast} and~\ref{Sec:Inference}, we demonstrate the applicability of these contrastive features in two applications: contrastive explanations and contrastive inference.

\section{Contrastive Explanations}
\label{Sec:Contrast}
Explanations are a set of rationales used to understand the reasons behind a decision~\cite{kitcher1962scientific}. In this section, we visually inspect the reasons behind decisions by answering \emph{'Why P, rather than Q?'} questions between the predicted class $P$ and the contrast class $Q$ for a network $f()$. We modify the popular Grad-CAM~\cite{selvaraju2017grad} explanation technique to obtain our contrastive visualizations. We first describe Grad-CAM before detailing the necessary modifications.
\vspace{-3mm}
\subsection{Grad-CAM}
\label{subsec:Gradcam}
Grad-CAM is used to visually justify the decision $P$ made by a classification network by answering \emph{`Why P?'}. The activations from the last convolutional layer of a network are used to create these visualizations since they possess high-level semantics while maintaining spatial information. For any class $i, \forall i \in [1,N]$, the logit $y_i$ is backpropagated to the feature map $A_l$ where $A_l = f_l(x)$ and $l$ is the last convolutional layer. The gradients at every feature map are $\frac{\partial y_i}{\partial A_l^k}$ for a channel $k$. These gradients are global average pooled to obtain importance scores $\alpha_k$ of every feature map in $l^{th}$ layer and $k^{th}$ channel. The individual maps $A_l^k$ are multiplied by their importance scores $\alpha_k$ and averaged to obtain a heat map. The Grad-CAM map at layer $l$ and class $i$ is given by $L^i = ReLU(\sum_{k=1}^K \alpha_k A^k_l )$. The Grad-CAM from an ImageNet-pretrained VGG-16~\cite{Simonyan15} for a correctly classified Spoonbill image is visualized in Fig.~\ref{fig:Contrast_examples}c. The red-highlighted regions in Fig.~\ref{fig:Contrast_examples}c explains why VGG-16 chose Spoonbill as the decision $P$. Hence, Grad-CAM visually explains the observed causality \emph{`Why P?'}.
\vspace{-3mm}
\subsection{Contrast Visualizations}\label{subsec:Contrast_Visuals}
In Grad-CAM, the importance score $\alpha_k$ which is derived by backpropagating the logit $y_i$, weighs the activations in a layer $l$ based on $A_l^k$'s contribution towards the classification. The activations are projections on network parameters and hence have access to both the causal and contrastive information. Therefore, to extract contrastive explanations, the contrast importance score $\alpha_k^c$ must be global average pooled contrastive features i.e $\alpha_k^c = \sum_u\sum_v \nabla_{W_l} J(W,x,P,Q)$, where $u,v$ are the channel sizes at layer $l$. This is achieved by backpropagating $J(W,x,P,Q)$ within the Grad-CAM framework to obtain the contrastive maps for class $Q$. Hence, while $\alpha_k$ highlights \emph{`Why P?'}, $\alpha_k^c$ denotes \emph{`Why P, rather than Q?'}. Note that there can be $N$ contrastive maps for a network trained to discriminate between $N$ classes. The contrast-emphasized regions for selected classes are shown in Fig.~\ref{fig:Contrast_examples}d-g. In Fig.~\ref{fig:Contrast_examples}d, VGG-16 indicates that the contrast between spoonbill and its notion of a flamingo class resides in the lack of S-shaped neck for a spoonbill. Similarly, it translates to not detecting white feathers in Fig.~\ref{fig:Contrast_examples}d to contrast between a spoonbill and a crane. The contrast between a band-aid and a spoonbill is in the presence of neck and legs in the spoonbill. This is highlighted in Fig.~\ref{fig:Contrast_examples}f. Fig.~\ref{fig:Contrast_examples}e indicates that VGG-16 contrasts between a pig and a spoonbill based on the neck of spoonbill. The body and feather colors of the spoonbill are de-emphasized but the shape of its legs and neck contribute towards VGG-16's decision. 

\begin{figure*}[!htb]
\begin{center}
\minipage{1\textwidth}%
  \includegraphics[width=\linewidth]{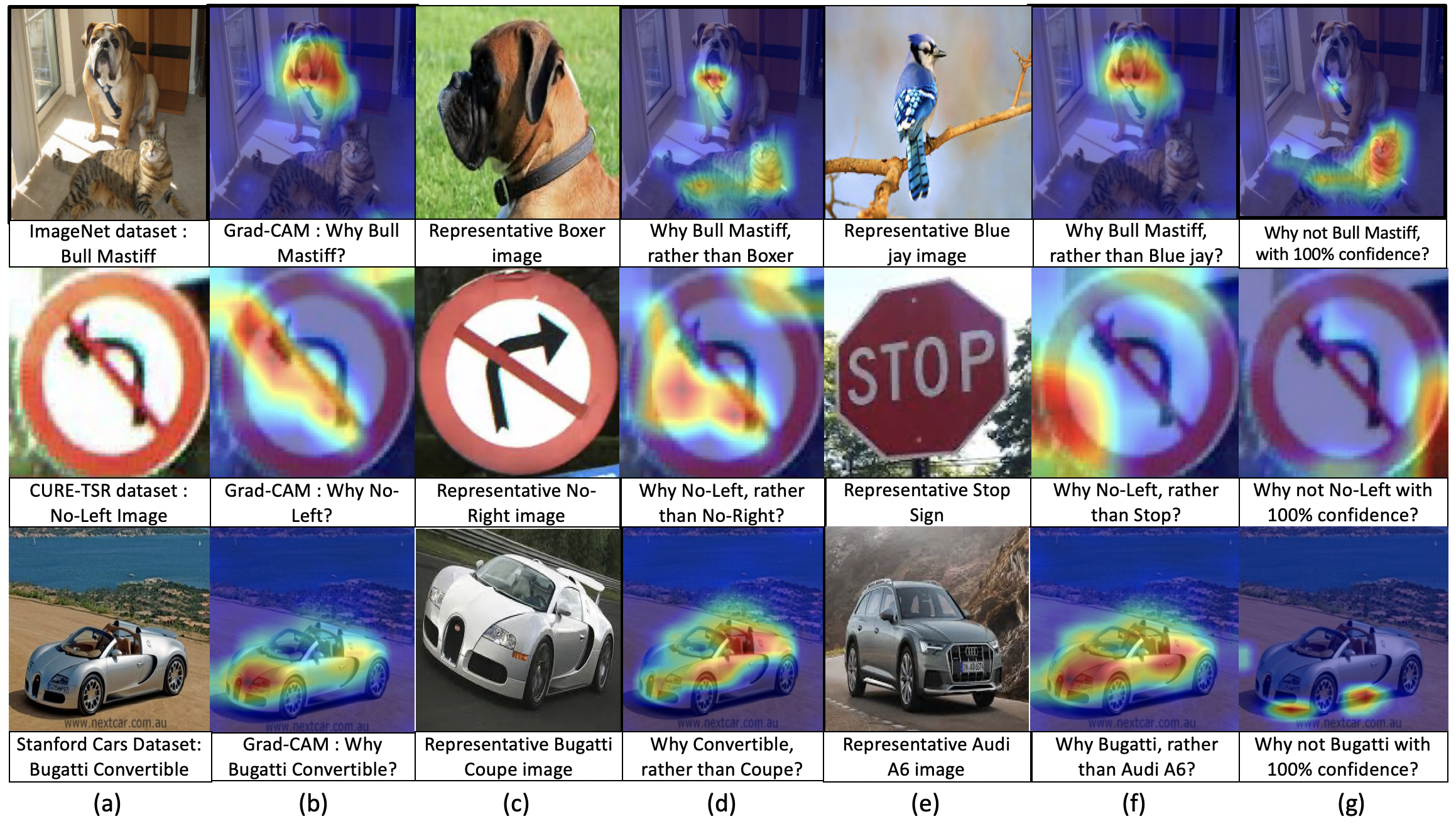}
\endminipage
\caption{Contrastive explanations (CE) on Recognition. (a) Input $x$. (b) Grad-CAM of $x$ for predicted class $P$. (c) Representative image of nearest class $Q$. (d) CE for class $Q$. (e) Representative image of random class $Q$. (f) CE for random class $Q$ in (e). (g) CE when $P = Q$.}\label{fig:Reasoning}
\end{center}
\end{figure*}

\subsection{Analysis}
\label{subsec:Grad-CAM_Analysis}
Consider an $L-$layered network $f()$ trained using the cross entropy loss loss $J()$ to differentiate between $N$ classes. The output of the network for an image $x$ after the last $L^{th}$ layer and before applying the loss is $y^L_{feat} = f_L(x)$ where $y^L_{feat}$ is a $1\times N$ vector containing the logits for each class. For brevity, we express $y^L_{feat}$ as $y$ in the remaining of this section. The general cross entropy loss for the label $i$ where $i \in [1,N]$ is,
\begin{equation}\label{eq:CE}
J(W, x, P, i) = -y_i + \sum_{j=1}^{N} e^{y_j}, \text{ where } y_j = f_L(x).
\end{equation}
During training, $i$ is the true label.

\noindent\textbf{Why P:} Note that in Grad-CAM the logit for the interested class is backpropagated. To ask \emph{`Why P?'}, the logit corresponding to class $P$, i.e $y_p$, is backpropagated. We represent the backpropagated variable in Grad-CAM as $J_G(P, P)$ where,
\begin{equation}\label{eq:Grad-CAM}
J_G(P, P) = y_P. 
\end{equation}
\noindent\textbf{Why P, rather than Q: }Contrast maps over $N$ classes are obtained by backpropagating the loss between predicted and contrast class $Q$. We represent this backpropagated variable as $J_C(P, Q)$ in Eq.~\ref{eq:CE}, where $Q \in [1,N]$ and $Q\neq P$. Approximating the exponent with its second order Taylor series expansion, we have,
\begin{equation}
J_C(P, Q) = -y_Q + \sum_{j=1}^{N} \bigg(1 + y_j + \frac{y_j^2}{2}\bigg).
\end{equation}
Note that for a well trained network $f()$, the logits of all but the predicted class are high. Hence $\sum_{j=1}^{N} \frac{y_j^2}{2} = \frac{y_p^2}{2}$. Substituting,
\begin{equation}\label{eq:J_Full}
J_C(P, Q) = -y_Q + N + \sum_{j=1}^N y_j + \frac{y_P^2}{2}.
\end{equation}
The quantity in Eq.~\ref{eq:J_Full} is differentiated, hence nulling the effect of constant $N$. For a well trained network $f()$, small changes in $W$ do not adversely affect the sum of all logits $\sum_{j=1}^N y_j$. Hence approximating its gradient to $0$ and discarding it, we can obtain $J_C(P, Q)$ as a function of two logits, $y_Q$ and $y_P$ given by,
\begin{equation}\label{eq:J_Final}
J_C(P, Q) = -y_Q + \frac{y_P^2}{2}.
\end{equation}
Compare Eq.~\ref{eq:J_Final} against Eq.~\ref{eq:Grad-CAM}. From Eq.~\ref{eq:Grad-CAM}, only $y_P$ or the logit for class $P$ is backpropagated to obtain importance scores. Hence, the importance score $\alpha_k$, highlights features in learned $l^{th}$ layer manifold where $f_l(x)$ projects onto patterns that justify $P$. In Eq.~\ref{eq:J_Final}, the backpropagated gradients are a function of $-y_Q$ and $y_P$. Hence, the contrast importance score $\alpha_k^c$, highlights the non-common features between classes $P$ and $Q$. These non-common features span the difference between $m_P$ and $m_Q$ from Fig.~\ref{fig:Contrast_examples}a. Recall $m_P$ is the learned manifold where $x$ is classified as $P$ and $m_Q$ is the hypothetical contrast manifold where $x$ is labeled as $Q$.

\noindent\textbf{Why P, rather than P: }When $Q = P$, Eq.~\ref{eq:J_Final} is written as,
\begin{equation}\label{eq:J_PP}
J_C(P, P) = -y_P + \frac{y_P^2}{2}.
\end{equation}
Note that the importance scores when backpropagating the first term in $J_C(P, P)$ are the negative of the backpropagated variable in Eq.~\ref{eq:Grad-CAM}. The authors in Grad-CAM~\cite{selvaraju2017grad} claim that backpropagating the negative score $-y_P$, provides conterfactual explanations. Since $J_C(P, P)$ is a function of both $y_P$ and $-y_P$, our results do not provide the results in the same counterfactual modality. However, since we are backpropagating a loss function between manifold $m_P$ against itself, the importance scores highlight those features in the image which disallow $f()$ to predict $x$ as $P$ with a higher confidence. If $J()$ were MSE, this modality reads as \emph{`Why not P with 100\% confidence?'}. 

\subsection{Qualitative Results}
\label{subsec:Explanations_Qual}
\subsubsection{Experiments}
In this section, we consider contrastive explanations on large scale and fine-grained recognition. Large-scale datasets, like ImageNet~\cite{ILSVRC15}, consist of a wide variety of classes. Fine-grained recognition is the subordinate categorization of similar objects such as different types of birds, and cars among themselves~\cite{yang2012unsupervised}. We consider Stanford Cars~\cite{KrauseStarkDengFei-Fei_3DRR2013} and traffic sign recognition CURE-TSR~\cite{Temel2017_CURETSR} datasets for fine-grained recognition. On ImageNet, we use PyTorch's ImageNet pretrained VGG-$16$~\cite{simonyan2014very} architecture to show results in Fig.~\ref{fig:Reasoning}. VGG-$16$ is chosen to be consistent with Grad-CAM. Note that we tested generating contrastive explanations on other architectures including AlexNet~\cite{krizhevsky2012imagenet}, SqueezeNet~\cite{iandola2016squeezenet}, VGG-$19$~\cite{simonyan2014very}, ResNet-$18,34,50,101$~\cite{he2016deep}, and DenseNet-$161$~\cite{huang2017densely}. On Stanford Cars dataset, we replace and train the final fully connected layer of an ImageNet pre-trained VGG-$16$ architecture to discriminate between $196$ classes. For CURE-TSR, we use the trained network provided by the authors in~\cite{Temel2017_CURETSR}. The results from the fine-grained datasets and the cat-dog image used in Grad-CAM~\cite{selvaraju2017grad} are shown in Fig.~\ref{fig:Reasoning}. Note that the time taken to generate a contrastive explanation is the same as Grad-CAM.

\begin{figure*}[!htb]
\begin{center}
\minipage{1\textwidth}%
  \includegraphics[width=\linewidth]{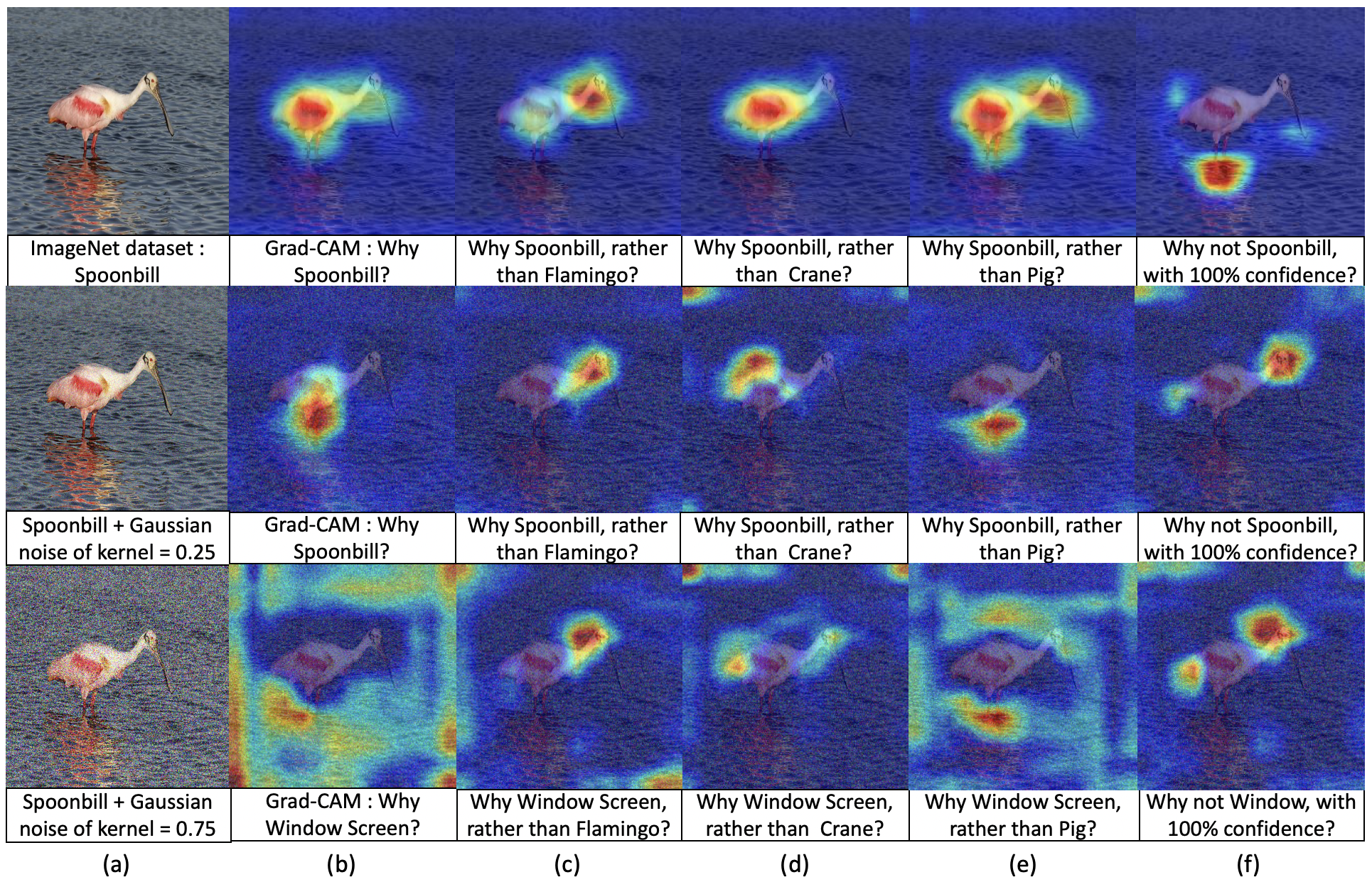}
\endminipage
\caption{Contrastive explanations (CE) on Recognition. (a) Input $x$. (b) Grad-CAM of $x$ for predicted class $P$. (c) Representative image of nearest class $Q$. (d) CE for class $Q$. (e) Representative image of random class $Q$. (f) CE for random class $Q$ in (e). (g) CE when $P = Q$.}\label{fig:Wrong_Reasoning}
\end{center}
\end{figure*}

\subsubsection{Results}
The visual answers to \emph{`Why P, rather than Q?'} from different datasets are provided in Fig.~\ref{fig:Reasoning}. We visualize the considered images $x$ in the column (a) of Fig.~\ref{fig:Reasoning}. Their respective Grad-CAM~\cite{selvaraju2017grad} explanations are shown in Fig~\ref{fig:Reasoning}b. Our contrastive explanations are visualized in columns (d), (f), and (g) of Fig.~\ref{fig:Reasoning}. The questions answered by each of the contrastive explanations is shown below the image. Representative images of the considered \emph{Q} class are shown alongside in columns (c) and (e) respectively. Note that the network does not see these representative images to produce the explanations. It bases its explanations on its notion of the \emph{Q} class. 

The contrastive explanations provide an interesting insight into the decisions of neural networks. For instance, the network's explanation as to \emph{`Why Bull Mastiff, rather than Boxer?'} in Fig.~\ref{fig:Reasoning}d is human interpretable. From the Boxer's representative image the two dogs differ in the shape of jowls. This is shown in the red highlights below the Bull Mastiff's jaws. Note that the red highlighted portion in the contrastive explanation is below the red highlighted region from Grad-CAM's explanation. This leads humans to interpret that the network's notion of the difference between the two breeds is the size and/or shape of jowls. However, contrastive explanations need not always be human interpretable. Consider the case when a network trained on a traffic sign dataset, CURE-TSR~\cite{Temel2017_CURETSR}, is asked \emph{`Why No-Left, rather than Stop?'}. The letters that spell STOP not being in the sign is the intuitive human response to the above question. However, from the contrastive explanation in Fig.~\ref{fig:Reasoning}f the network highlights the bottom left corner of the image. Among the 14 traffic signs the network has trained to differentiate between in the dataset, Stop sign is the only class that has a hexagonal shape. Hence, the network has learned to check for a straight side in the bottom left. The absence of this side in $x$ indicates to the network that $x$ is not a STOP sign. This clearly illustrates the differences between the notion of classes between humans and machines. When the difference between $P$ and $Q$ are not fine-grained, then the contrastive explanations are similar to Grad-CAM explanations. This is illustrtaed by showing explanations between predicted $P$ and random classes in Fig.~\ref{fig:Reasoning}f for ImageNet and Stanford Cars dataset. The difference between a dog and a bluejay is in the face of the dog - the same region highlighted by Grad-CAM to make decisions. The input Bugatti Veyron's sloping hood is sufficiently different from that of the boxy hood of the Audi that it is highlighted. The same region is used to classify it as a Bugatti. Fig.~\ref{fig:Reasoning}g provides contrastive explanations when $P$ is backpropagated. We use the question \emph{`Why not P with 100\% confidence?'} as the explanation modality. However, as noted before, this question is loss dependent. The results in cat-dog image are human interpretable - the presence of the cat prevents the network from making a confidant decision. 
\vspace{-0mm}
\subsubsection{Results on noisy data}\label{subsubsec:Noisy_explanations}
$P$ is the network prediction and hence is not controlled in the statement \emph{`Why P, rather than Q?'}. In this section, we add noise to $x$ to illustrate the effect of noise as well as incorrect classification $P$ for contrastive explanations. The results are presented in Fig.~\ref{fig:Wrong_Reasoning}. The first row shows the pristine image of spoonbill along with Grad-CAM and contrastive explanations. The second row illustrates the scenario when the spoonbill image has Gaussian noise added to it. This noise however, is insufficient to change prediction $P$. In the third row, the noisy spoonbill image is classified as a window screen. In all three rows, both the Grad-CAM and contrastive explanations change. We first analyze Grad-CAM. For the pristine spoonbill image, the network infers based on the body of the spoonbill. When small noise is added, the correct classification is made based primarily on the legs. When a large amount of noise is added, the network incorrectly predicts the spoonbill as a window screen based on features around the bird. Among the contrastive explanations for why the image is not a flamingo, the network consistently highlights the neck. The results for crane also highlight the body. The results for \emph{`Why not window with 100\% confidence?'} highlights the face and tail of the bird which is intuitive. Hence, contrastive explanations provide additional context and information that is not available from observed causal explanations. We propose to tie inference to contrastive patterns rather than associative feed-forward patterns so as to obtain additional features to infer from. In other words, not only does the bird image have to have the requisite patterns for a spoonbill, it also has to satisfy the \emph{`Why not Q?'} modality where $Q\in [1,N]$. Inference based on all possible contrastive features is contrastive inference. In Section~\ref{Sec:Inference}, we illustrate the robustness of contrastive inference.
\vspace{-3mm}

\begin{figure*}[!htb]
\begin{center}
\minipage{1\textwidth}%
  \includegraphics[width=\linewidth]{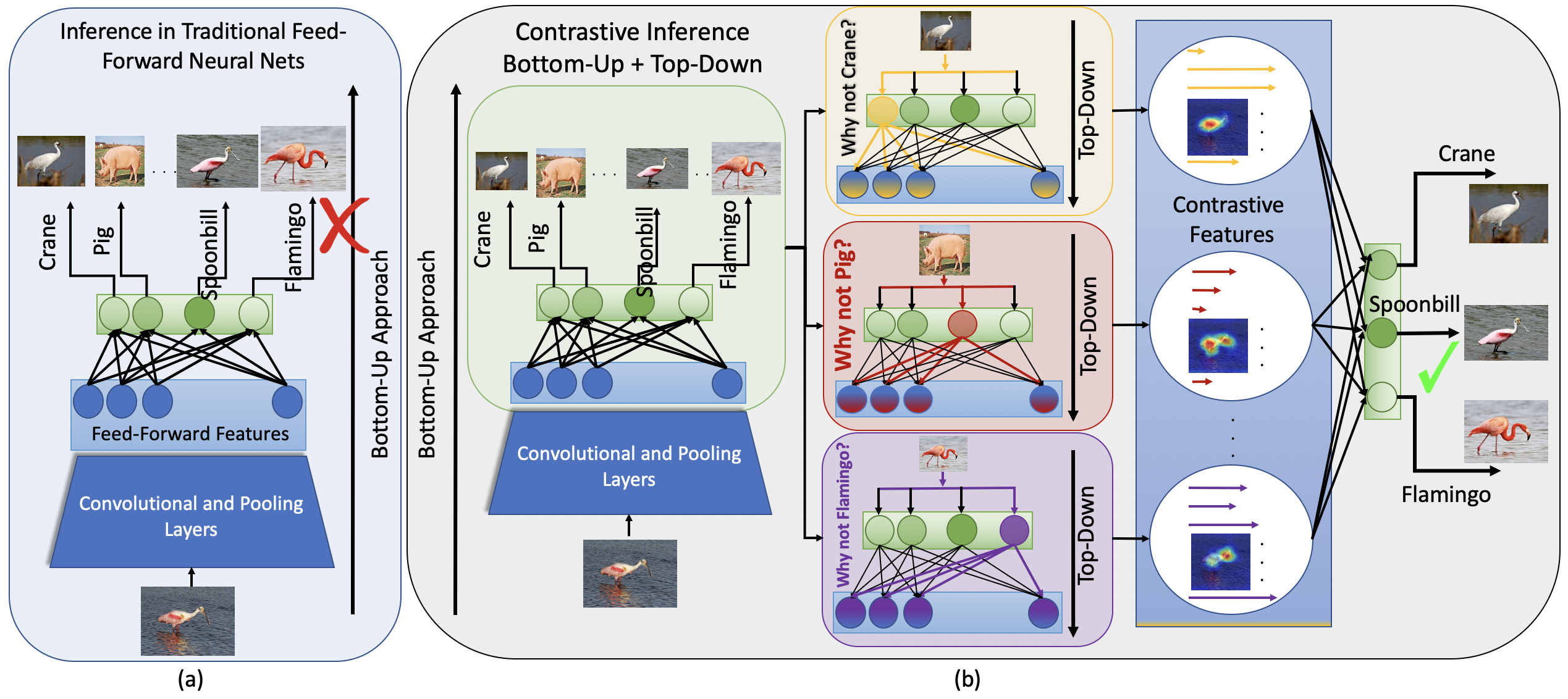}
\endminipage
\caption{Feed-Forward Inference vs Contrastive Inference. (a) The features in blue before the final fully connected layer represent the feed-forward features. (b) The features for contrastive inference represent the change in a learned model when given a contrast hypothesis.}\label{fig:Contrastive_Inference}\vspace{-3mm}
\end{center}
\end{figure*}
\section{Contrastive Inference}
\label{Sec:Inference}
In Section~\ref{Sec:LitReview}, we stated that abductive reasoning is more robust to new conditions and stimuli. In this section, we use the obtained contrastive features to validate this statement. Contrastive reasons, like abductive reasons, generalize under new conditions. We train on pristine images and test on noisy images to verify robustness. Neural networks have shown a vulnerability to distortions between train and test domains~\cite{Temel2017_CURETSR}. We first describe the feed-forward inference from existing neural networks that use inductive reasoning and feed-forward features from Section.~\ref{subsec:FF-Features} to make decisions. This is followed by our proposed contrastive inference framework. 

\subsection{Feed-Forward Inference}\label{subsec:FF-Inference}
In Section~\ref{subsec:FF-Features}, we described the activations that form the feed-forward features that result in inductive decisions. Continuing the notations used in Section~\ref{subsec:FF-Features}, we consider the inference mechanism for the specific task of classification. For a network $f()$ that is trained to classify between $N$ classes, the last layer is commonly a fully connected layer consisting of $N$ weights or filters. During inference the feed-forward features at the $(L-1)^{\text{th}}$ layer, $y_{feat}^{L-1} = f_{L-1}(x)$, are projected independently onto each of the $N$ filters. The filter with the maximum projection is inferred as the class to which $x$ belongs to. Mathematically, feed-forward features $y_{feat}^l$ and the network prediction $P$ are related as,
\begin{align}
    y_{feat}^l = & f_{l}(x), \forall l \in [1, L-1],\\
    P = & \operatorname*{arg\,max} (W_L^T y_{feat}^{L-1} + b_L), \label{eq:FF-Infer}\\
    \forall  W_L\in \Re^{d_{L-1}\times N}, & y_{feat}\in \Re^{d_{L-1}\times 1}, b_L\in \Re^{N\times 1}, 
\end{align}
where $W_L$ and $b_L$ are the parameters of the final linear layer. 
\vspace{-3mm}
\subsection{Contrastive Inference}
For the application of classification, the workflow is shown in Fig~\ref{fig:Contrastive_Inference}. For a pretrained neural network trained on $N$ classes on $X$ domain, an image $z \in Z$ such that $Z \neq X$ is provided. For $f()$ that is not adapted to test domain $Z$, the prediction is incorrect as illustrated in Fig~\ref{fig:Contrastive_Inference}. We hypothesize that the actual prediction should be all possible $N$ classes. By feeding in the new hypothesis $Q$, we obtain contrastive features from $f()$. We represent the data $x$ as a measure of change that it creates in the trained base network against all classes. These features are used to obtain the correct prediction. We first expand on the data representation part before obtaining predictions from them.

\subsubsection{Contrastive Data Representation}
For discriminative networks, contrast for class $1$ is provided by backpropagating class $1$. The gradient is proportional to the loss function $J(W,x,P,1)$, where $W$ is the weight parameter and $J()$ is a convex loss function. Specifically, the manifold change is in the direction of $\nabla_{W_l} J(W,x,P,1)$ for weights in layer $l\in [1,L]$. This term is the gradient of loss for $1$ with respect to weights in layer $l$. As shown in Fig~\ref{fig:Contrastive_Inference}b, we backpropagate over all $N$ classes to obtain contrastive features across all classes given by, $r_i = \nabla_{W_l} J(W,x,P,i)$. The final contrastive feature, $r_x$ for an image $x$, is given by concatenating all individual contrasts. Hence,
\begin{equation}\label{eq:r_dis}
\begin{gathered}
    r_i = (\nabla_{W_l} J(W,x,P,i)), \forall l \in [1,L], \forall i \in [1,N] \\
    r_x = [r_1, r_2 \dots r_N]
\end{gathered}
\end{equation}
The loss function $J$ is taken with respect to the logits after the final layer. This loss is representative of the contrast between the predicted output $P = f(x)$ and the contrast, $i \in [1,N]$. Loss $J()$ need not be the same loss used to train the network. In Section~\ref{Sec:Contrast}, we used cross entropy to visualize explanations. Access to quantitative results allows for testing other loss functions during inference. In Section~\ref{sec:Results}, we test a number of available loss functions based on classification accuracy. In this work, we use a modified MSE loss function where $J = \text{MSE}(f_L(x), \delta^M_i)$  where $\delta^M_i$ is a modified Kronecker delta function given by,
\begin{equation}\label{eq:Kroenecker}
    \delta^{M}_{i} =
    \begin{cases}
            M, &         \text{if } i=\text{class},\\
            0, &         \text{otherwise} 
    \end{cases}
\end{equation}
where $M$ is the mean of output logits $y_P = \text{max }f(x)$ taken for all $x$ in training set. We use $M$ instead of $1$ because we want the network to be as confidant of the contrast as it is with the prediction and penalize it accordingly. Note that we now have our contrastive features $r_x$ for a datapoint $x$.

\subsubsection{Inference Based on Contrastive Data Representation}
Once $r_x$ is obtained for all $N$ classes, the contrastive feature is now analogous to $y_{feat}^{L-1}$ from Eq.~\ref{eq:FF-Infer}. Similar to $y_{feat}$, $r_x$ is processed as Eq.~\ref{eq:FF-Infer}. However, $y_{feat}$ is of dimension $\Re^{d_{L-1}\times 1}$ and $r_x$ is of dimension $\Re^{(N\times d_{L-1})\times 1}$ since it is a concatenation of $N$ gradients. To account for the larger dimension size, we classify $r_x$ by training a simple Multi Layer Perceptron (MLP), that we will henceforth refer to as $\mathcal{H}$, on top of $r_x$ derived from training data. The structure of the MLP depends on the size $(N \times d_{L-1}) \times 1$. This is provided in the implementation details in Section~\ref{Subsec:In-Distribution}. Once $r_x$ passes through $\mathcal{H}$, the argument of the maximum logit $\Tilde{y}$, is inferred as the class of $x$. Expressing $\Tilde{y}$ mathematically, we have,
\begin{align}
r_x = [r_1, r_2, \dots r_N]&, \forall r_i = (\nabla_{W_l} J(W,x,P,i)),\label{eq:CI-features}\\
\Tilde{y} &= \operatorname*{arg\,max} (\mathcal{H}(r_x)).\label{eq:CI-Inrer}
\end{align}
Notice the similarity between the feed-forward prediction $P$ and contrastive prediction $\Tilde{y}$. Our goal is to make both the feed-forward and contrastive workflows similar while only changing the feed-forward features to contrastive features so as to showcase the effectiveness of contrastive reasoning during inference.

Since the proposed contrastive inference follows an abductive reasoning approach, we borrow the mathematical interpretation of abductive reasoning to formulate contrastive inference. The authors in~\cite{douven2015probabilistic} suggest that abductive reasoning is a non-bayesian process and provide an update rule for training. Adapting this to the inference stage, we have, 
\begin{equation}\label{eq:probability}
    \Tilde{y} = \argmax_i{\mathcal{F} \big[ P, {\mathcal{H}}(Pr(i|P), r_x)\big]},
\end{equation}
where $\Tilde{y}$ is the prediction from contrastive inference from Eq.~\ref{eq:CI-Inrer}, $P$ is the prediction of feed-forward inference from Eq.~\ref{eq:FF-Infer}, $Pr(i|P)$ is the probability that the true prediction is the contrast class $i \in [1,N]$ conditioned on the given feed-forward prediction $P$, and $r_x$ is the contrastive feature. Consider Fig~\ref{fig:Contrastive_Inference}. Given the image of a spoonbill $x$, let the feed-forward inference predict $P$ as a flamingo. $r_x$ is extracted from $f()$ which is used to train another classifier that represents ${\mathcal{H}}$ from Eq.~\ref{eq:probability}. The entire contrastive inference framework requires both feed-forward prediction and ${\mathcal{H}}$. This framework is represented by the function ${\mathcal{F}}$ from Eq.~\ref{eq:probability}. ${\mathcal{F}}$ is the proposed contrastive inference.

While feed-forward inference obeys Bayes rule to obtain $P$, Eq~\ref{eq:probability} is non-Bayesian. It depends not only on the learned network $f()$, but also on contrastive features $r_x$ derived from class hypotheses - features that represent effect to cause. However, in the limiting case when a network is well trained, Eq.~\ref{eq:probability} behaves in a Bayesian fashion. This is because of the $Pr(i|P)$ term. In an ideal case, for a well trained network $f()$ with train and test data from the same distribution, $Pr(i|P) = 1$ for $i = P$ and $0$ otherwise. We test this hypothesis in Table~\ref{table:Original_Results} when networks $f()$ trained and tested on distribution $X$, perform with the same accuracy in both feed-forward and contrastive settings.
\vspace{-3mm}
\begin{table*}
\centering
\small{}
\caption{Structure of $\mathcal{H}$. R refers to ResNet.}
\vspace{1mm} 
\begin{tabular}{c c c c c } 
 \hline
 Section & Train Data & $f()$ & $\mathcal{H}$ - All layers separated by sigmoid & Time To Train (sec) \\
 \hline\hline
 \ref{Subsec:In-Distribution}, \ref{Subsec:Distortion} & CIFAR- & R-18,34 &  $640\times300-300\times100-100\times10$ & 60\\ 
 \ref{subsec:f-Effect}, \ref{subsec:f-Effect}   & 10   & R-50,101 & $2560\text{x}1000-1000\text{x}500-500\text{x}300-300 \text{x}100-100\text{x}10$ & 300\\ 
 \hline
 \ref{subsubsec:DA} & VisDA & R-18 & $768\times300-300\times100-100\times12$ & 67\\
 \hline
\end{tabular}
\label{table:Structure}\vspace{-3mm}
\end{table*}

\section{Results}
\label{sec:Results}

In Section~\ref{Sec:Contrast}, the contrastive reasons were used to demonstrate the visual explanations that can complement existing \emph{`Why P?'} explanations. In this section, we demonstrate the robustness claim made in Section~\ref{Sec:Inference}. We apply feed-forward and contrastive inference on existing neural networks and show that : 1) contrastive inference performs similarly to feed-forward inference when $f()$ is trained and tested on pristine data, 2) contrastive inference outperforms its feed-forward counterpart when the testing data is distorted. Distortions include image acquisition errors, severe environmental conditions during acquisition, transmission and storage errors among others. Current techniques that alleviate the drop in feed-forward accuracy require training using distorted data. The authors in~\cite{vasiljevic2016examining} show that finetuning or retraining networks using distorted images increases the performance of classification under the same distortion. However, performance between different distortions does not generalize well. For instance, training on Gaussian blurred images does not guarantee a performance increase in motion blur images~\cite{vasiljevic2016examining, geirhos2018generalisation}. Other proposed methods include training on style-transferred images~\cite{geirhos2018imagenet}, training on adversarial images~\cite{hendrycks2019benchmarking}, and training on simulated noisy virtual images~\cite{Temel2017_CURETSR}. All these works require additional training data not belonging to $X$. In this paper, we show that contrastive inference increases classification accuracy in $19$ considered distortions from CIFAR-10-C dataset~\cite{hendrycks2019benchmarking}. This increase in accuracy is induced because of the contrastive features extracted from the pristine data and not through training with distorted data.
\vspace{-3mm}
\subsection{Results on Pristine Data}
\label{Subsec:In-Distribution}
We train four networks - ResNet-18,34,50, and 101~\cite{he2016deep} on CIFAR-10~\cite{krizhevsky2009learning} trainset and test them on CIFAR-10 testset. The training set of CIFAR-10 has $50000$ images from $10$ classes with each class having $5000$ sample images. The networks are trained in PyTorch for $200$ epochs on a NVIDIA 1080Ti GPU with a batch size of $128$ using SGD optimization. Learning rates of $0.1, 0.004,$ and $0.0008$ are used from epochs $0-60, 60-120,\text{ and } 160-200$ respectively along with a momentum of $0.9$ throughout the training procedure. PyTorch's horizontal flipping transformation is used as a data augmentation technique. The testset in CIFAR-10 consists of $10000$ images with each class represented by $1000$ images. The results of all networks derived using Eq.~\ref{eq:FF-Infer} are shown as feed-forward results in Table~\ref{table:Original_Results}. 
\begin{table}[h!]
\small
\centering
\caption{Feed-Forward Inference vs contrastive Inference on CIFAR-$10$ test set}
\vspace{-1mm}
\begin{tabular}{c | c c c c} 
 \hline
  ResNet & 18 & 34 & 50 & 101 \\ [0.5ex] 
 \hline\hline
 Feed-Forward (\%) & $91.02$ & $93.01$ & $93.09$ & $93.11$ \\ [1ex] 
 \hline
 Gradients (\%) & $90.94$ & $93.14$ & $92.88$ & $92.73$ \\ [1ex] 
 \hline
\end{tabular}
\label{table:Original_Results}\vspace{-7mm}
\end{table}
For all $50000$ images in the training set, we extract the contrastive features $r_x$. $r_x$ is richer in terms of dimensionality and contrastive information. For instance, for every image in ResNet-18, the feed-forward network provides a $64\times 1$ feature as $y^{L-1}_{feat}$ from Eq.~\ref{eq:FF-Infer}. Using these feed-forward features, the proposed contrastive features, $r_x$ from Eq.~\ref{eq:r_dis}, are extracted with a dimensionality of $640\times1$ feature. These features are normalized. $\mathcal{H}$, with the structure given in Table~\ref{table:Structure} is trained on the $50000$ training contrastive features for $200$ epochs using the same procedure as the original network. The time to train $\mathcal{H}$ is dependent on its structure and is shown in Table~\ref{table:Structure}. Hence, for contrastive inference, our full framework consists of the original trained network, $f()$, and an MLP, $\mathcal{H}$.

During testing, all $10000$ images from the testset are passed through the original network. The contrastive features from each image are extracted individually. These features are normalized and passed through the trained MLP. The overall accuracy is reported in Table~\ref{table:Original_Results}. As can be seen, the results of all three methods are comparable to that of the feed-forward procedure. These results are a validation of the Bayesian limiting case scenario expressed in Eq.~\ref{eq:probability}. When the network $f()$ is well trained from a distribution $X$, $Pr(i|P,x\in X)$ where $i\in [1,N]$ provides no new information regarding any $i$ other than $P$. Hence, the results from Eq.~\ref{eq:probability} are only influenced by the $64\times 1$ feature extracted using $i = P$ in Eq.~\ref{eq:FF-Infer}.
\vspace{-3mm}
\subsection{Results on Distorted Data}
\label{Subsec:Distortion}

\begin{figure*}[!htb]
\begin{center}
\minipage{\textwidth}%
  \includegraphics[width=\linewidth]{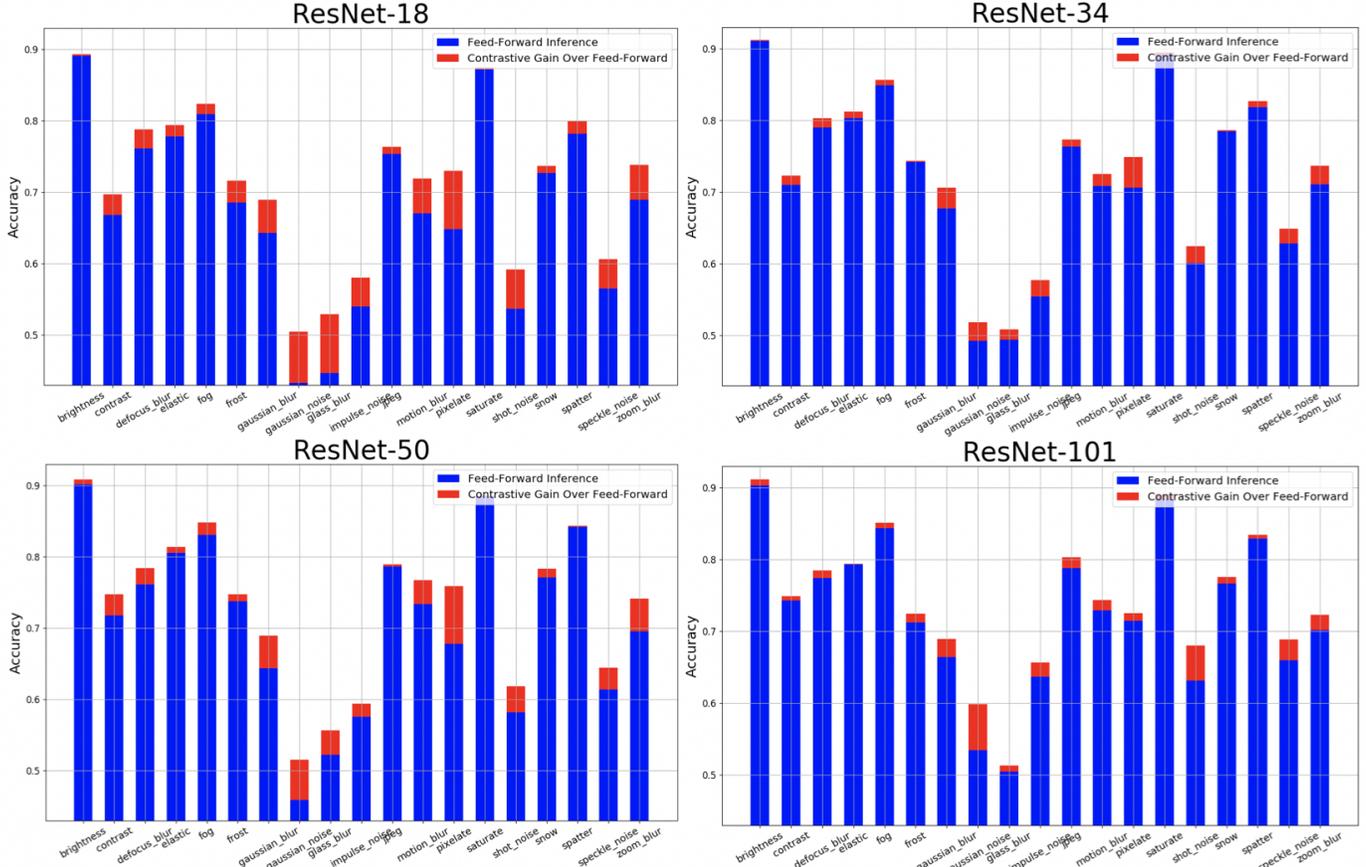}
\endminipage
\vspace{-3mm}
\caption{Visualization of accuracy gains (in red) of using the proposed contrastive inference over feed-forward inference on CIFAR-10-C for four networks among $19$ distortions. Within each distortion, the distortion levels are averaged and the results shown.}\vspace{-0.7cm}\label{fig:Robustness_Distortionwise}
\end{center}
\end{figure*}

\begin{figure*}[!htb]
\begin{center}
\minipage{0.49\textwidth}%
  \includegraphics[width=\linewidth]{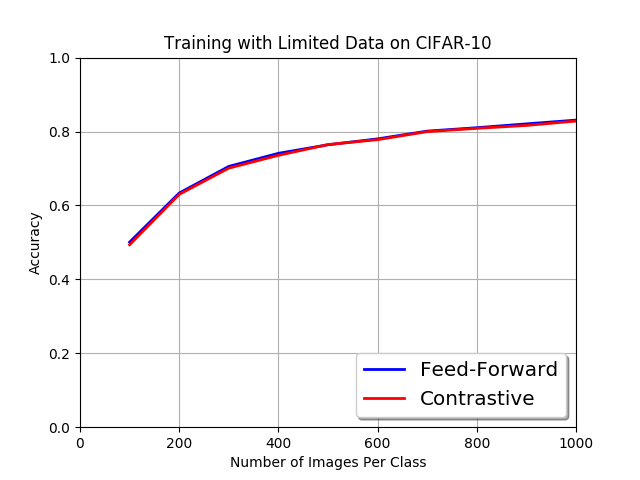}
\endminipage
\minipage{0.49\textwidth}%
  \includegraphics[width=\linewidth]{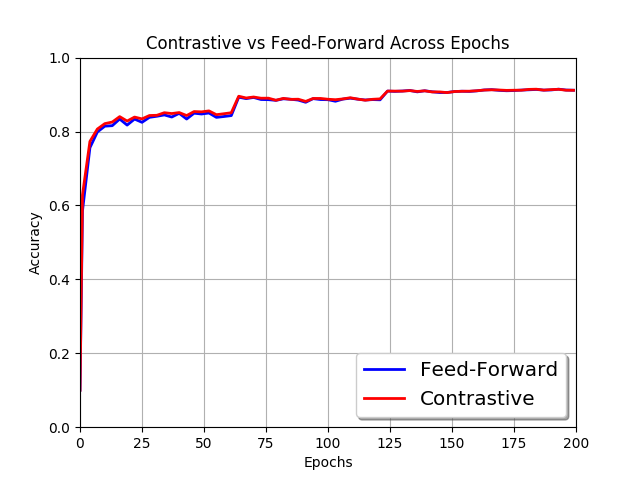}
\endminipage
\caption{(a) Contrastive inference vs. Feed-Forward Inference when $f()$ is trained on limited data. (b) Contrastive inference vs. Feed-Forward Inference when $f()$ is trained in limited time.}\label{fig:f-effect}\vspace{-0.7cm}
\end{center}
\end{figure*}

\begin{figure*}[!htb]
\begin{center}
\minipage{0.9\textwidth}%
  \includegraphics[width=\linewidth]{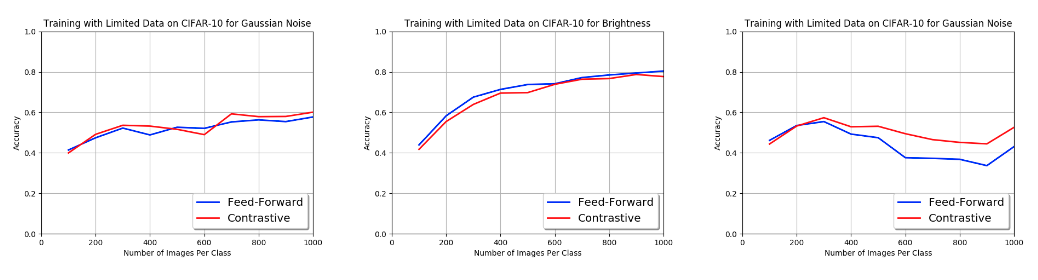}
\endminipage
\caption{Contrastive Inference vs. Feed-Forward Inference under limited training data on (a) Brightness distortion, (b) Motion blur distortion, (c) Gaussian noise distortion.}\label{fig:Data-Tests}\vspace{-0.6cm}
\end{center}
\end{figure*}

\begin{figure*}[!htb]
\begin{center}
\minipage{0.9\textwidth}%
  \includegraphics[width=\linewidth]{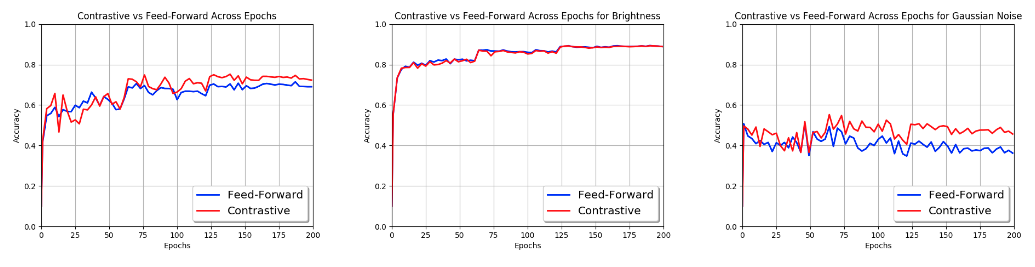}
\endminipage
\caption{Contrastive Inference vs. Feed-Forward Inference across epochs under (a) Brightness distortion, (b) Motion blur distortion, (c) Gaussian noise distortion.}\label{fig:Epoch-Tests}\vspace{-0.7cm}
\end{center}
\end{figure*}

In this section, we use CIFAR-10-C~\cite{hendrycks2019benchmarking} to validate the performance of contrastive inference on distorted data. CIFAR-10-C provides $19$ distortions on the testset of CIFAR-10. It also provides $5$ increasing levels of degradation in each distortion. Hence, every distortion has $50000$ images to test on. $19$ distortions produce $950,000$ images during testing on a network trained on the original $50000$ images. The same networks ResNet-18,34,50,101 trained in Section~\ref{Subsec:In-Distribution} are used for the experiments in this section. The results are shown in Fig.~\ref{fig:Robustness_Distortionwise}. The blue bars depict the results when the network $f()$ infers the predictions of the distorted images in a feed-forward fashion. The red bar depicts the contrastive accuracy gain over the feed-forward predictions with the inclusion of $\mathcal{H}$. The results of all $19$ distortions averaged over $5$ distortion levels are shown. Among $4$ networks and in every distortion category, there is an increase in the performance of contrastive inference over feed-forward inference. However, the increase is not consistent across the distortion categories. The results of all $4$ networks increase substantially by atleast $5\%$ in $8$ of the $19$ categories. These distortions include - gaussian blur, gaussian noise, glass blur, motion blur, pixelate, shot noise, speckle noise, and zoom blur. The highest increase is $8.22\%$ on glass blur for ResNet-18. These distortions are global distortions i.e. the criteria and methodology for distorting pixels does not change based on individual pixels. Other distortions including contrast, and brightness where the proposed method's performance gains are less than $3\%$ consist of distortions that change the global and local characteristics of the image including it's mean and contrast. Neural networks are actively trained to ignore such changes so that their effects are not propagated beyond the first few layers. The contrast obtained from the last layer is by itself insufficient to characterize the underlying data without the noise. This should be addressed by deriving contrast from earlier layers but is currently beyond the scope of this paper. Also, the contrastive performance gain increases as the level of distortions increase. This is because, with higher distortion, we move further away from the Bayesian limiting case in Eq.~\ref{eq:probability}. In other words, the information in $Pr(i|P,x\in X)$ increases. For ResNet-18, the average level $1$ contrastive gain across all $19$ distortions is $1.36\%$ while its level $5$ gain across distortions is $5.26\%$.

\begin{table*}
\centering
\caption{Contrastive Inference averaged accuracies with different loss functions for ResNet-18 on CIFAR-10-C.}\vspace{-1mm}
\begin{tabular}{c c c c c c c c c c} 
 \hline
   Feed-Forward & MSE & CE & BCE & L1 & L1-M & Smooth L1 & Smooth L1-M & NLL & SoftMargin  \\ [0.5ex] 
 \hline\hline
   $67.89\%$ & $71.35\%$ & $69.42\%$ & $70.24\%$ & $69.09\%$ & $69.92\%$ & $69.22\%$ & $70.01\%$ & $70.93\%$ & $70.91\%$ \\ [1ex] 
 \hline\vspace{-7mm}
\end{tabular}
\label{table:Losses}
\end{table*}
\vspace{-3mm}
\subsubsection{Effect of loss in gradient generation}
Note that the gradient generation process is stochastic in Eq.~\ref{eq:CI-features}. We test the impact of the choice of loss functions in Eq.~\ref{eq:CI-features} on the performance of contrastive inference. We compare multiple loss functions all of whose distortion-wise level-wise averaged results are provided in Table~\ref{table:Losses}. Feed-forward accuracy is the results obtained from the network $f()$, MSE is the modified Mean Squared Error from Eq.~\ref{eq:Kroenecker}, CE is Cross Entropy, BCE is Binary Cross Entropy, L1 is Manhattan distance, L1-M is the modified version of L1 distance when $\delta^M_i$ from Eq.~\ref{eq:Kroenecker} is backpropagated, Smooth L1 is the leaky extension of Manhattan distance, Smooth L1-M is the modified version of Smooth L1 similar to L1-M, NLL is Negative Log Likelihood loss, and SmoothMargin is the modified hinge loss. All these loss functions are taken from PyTorch. In Table~\ref{table:Losses}, the performance of all contrastive inference modalities with different loss functions, exceed that of the feed-forward inference. The proposed modified MSE outperforms the nearest loss by $0.42\%$ in accuracy and is used in all our experiments.

\vspace{-3mm}
\subsection{Effect of $f()$}\label{subsec:f-Effect}
We further analyze ResNet-18 trained on CIFAR-10 to see the efficacy of contrastive features when $f()$ is \emph{not} well trained i.e. when contrast has not yet been learned implicitly. In this subsection, our goal is to ascertain that the performance gain obtained by contrast is not due to a poorly trained $f()$ or a statistical anomaly. We consider two cases of such $f()$ : when $f()$ is trained for limited time, and when $f()$ is trained with limited data. ResNet-18 is trained on CIFAR-10 with the same learning parameter setup as in Section~\ref{Subsec:In-Distribution} and Table~\ref{table:Original_Results}. 

\subsubsection{Training and testing under limited time}\label{subsubsec:LWLT}
For the limited time experimental setup, ResNet-18 is trained for $200$ epochs. The network states at multiples of $3$ epochs from $1$ to $198$ along with epoch $= 200$ are saved. This provides $67$ versions of $f()$ under different stages of training. Each $f()$ is tested on $10,000$ CIFAR-10 testing images and the Top-$1$ accuracy is plotted in blue in Fig.~\ref{fig:f-effect}b. The gradients $r_x$ for all $67$ states are extracted for the $50,000$ training samples. These gradients are used to train $67$ separate $\mathcal{H}$ of structure provided in Table~\ref{table:Structure} with a similar parameter setup as before. The gradients from the $10,000$ testing samples are extracted separately for each of the $67$ ResNets and tested. The results are plotted in red in Fig.~\ref{fig:f-effect}b. Note the sharp spikes at epochs $60$ and $120$ where there is a drop in the learning rate. Hence, when training domain is the same as testing domain, contrastive and feed-forward features provide statistically similar performance across varying states of $f()$ in accordance with Eq.~\ref{eq:probability}.

We now consider the case when a network $f()$ is trained for limited epochs on domain $X$ and tested on $Z$ from CIFAR-10-C distortions. The $67$ trained models of ResNet-18 are are tested on three distortions from CIFAR-10-C. The three distortions include motion blur, brightness, and Gaussian noise. These three distortion types were chosen to represent the spectrum of the proposed method's performance increase over feed-forward inference. From the results in Fig.~\ref{fig:Robustness_Distortionwise}, contrastive inference achieved one of its highest performance gains in Gaussian noise, lowest performance gain in brightness and an average increase in motion blur. The results in Fig.~\ref{fig:Epoch-Tests} indicate that after around $60$ epochs, the feed-forward network has implicitly learned contrasts sufficiently to discriminate between classes. This is seen in both motion blur and Gaussian noise experiments. The results from brightness indicate that contrastive inference follows feed-forward inference across epochs.

\subsubsection{Training and testing with limited data}\label{subsubsec:LimitedData}
For the limited data experiment, ResNet-18 is trained on only a subset of available data. CIFAR-10 consists of $50,000$ training images with each class consisting of $5,000$ images. We randomly sample a fixed number images from each class and train $f()$ using these samples. The results are plotted in Fig.~\ref{fig:f-effect}a. Ten separate ResNets are trained with each model having access to random $100$, $200$, $300$, $400$, $500$, $600$, $700$, $800$, $900$, and $1000$ images per class. Validation is conducted on all $10,000$ testing images and the results are plotted in blue in Fig.~\ref{fig:f-effect}a. Contrastive features are extracted for the same random images that the base $f()$ was trained on and $10$ separate $\mathcal{H}$ are trained on these gradients. $r_z$ for $10,000$ testing data is separately extracted for all ten instances of $f()$ and passed through trained $\mathcal{H}$ to obtain contrastive results. These are plotted in red. Similar to the results in Table~\ref{table:Original_Results}, contrastive inference is statistically similar to feed-forward inference. 

We consider the case when a network $f()$ is trained on limited data on $X$ but tested on distorted data $Z$, from CIFAR-10-C. We consider the same distortions as before from Section~\ref{subsubsec:LWLT}. The results of feed-forward and contrastive inference are plotted in Fig.~\ref{fig:Data-Tests}. For Gaussian noise distortion, contrastive inference outperforms feed-forward inference even with only $300$ training images per class. However, this is not the case for both motion blur and brightness where contrastive inference follows feed-forward inference.

\begin{table*}
\centering
\caption{Performance of Contrastive Inference vs Feed-Forward Inference on VisDA Dataset}
\vspace{-1.5mm} 
\begin{tabular}{c c c c c c c c c c c c c c} 
 \hline
    & Plane & Cycle & Bus & Car & Horse & Knife & Bike & Person & Plant & Skate & Train & Truck & All \\ [0.5ex] 
 \hline
 Feed-Forward & $27.6$ & $7.2$ & $\textbf{38.1}$ & $54.8$ & $43.3$ & $\textbf{4.2}$ & $\textbf{72.7}$ & $\textbf{8.3}$ & $28.7$ &  $22.5$ & $\textbf{87.2}$ & $2.9$ & $38.1$\\ 
 
 Contrastive & $\textbf{39.9}$ & $\textbf{27.6}$ & $19.6$ & $\textbf{79.9}$ & $\textbf{73.5}$ & $2.7$ & $46.6$ & $6.5$ & $\textbf{43.8}$ &  $\textbf{30}$ & $73.6$ & $\textbf{4.3}$ & $\textbf{43.6}$\\
 \hline
\end{tabular}
\label{table:VisDA}
\end{table*}
\vspace{-3mm}
\subsection{Effect of training data}\label{subsec:D-Effect}
We analyze the impact of training data in three additional cases : a) The base network $f()$ is trained on noisy images, b) the training and testing data are of a higher resolution than the $32\times 32 \times 3$ CIFAR-10 images, c) the training and testing data are significantly different like in VisDA dataset~\cite{peng2017visda}. In all three cases, we \emph{do not} use data from training domain to learn $\mathcal{H}$.
\vspace{-3mm}
\subsubsection{Results on training with noisy data}\label{subsubsec:NoiseTrain}
In Sections~\ref{Subsec:Distortion} and \ref{subsec:f-Effect}, $\mathcal{H}$ is used as a plug-in on top of neural network $f()$ trained on pristine data $X$. Consider a network $f'()$ that has trained on distorted data. We apply contrastive inference on $f'()$ and show that there is a performance gain when using $\mathcal{H}$. In this experimental setup, we augment the training data of CIFAR-10 with six distortions - gaussian blur, salt and pepper, gaussian noise, overexposure, motion blur, and underexposure - to train a ResNet-18 network $f'()$. We then test the network on CIFAR-10 test data corrupted by the same six distortions in $5$ progressively varying degrees. The distortions were obtained from~\cite{temel2018cure}. Note that while $f'()$ is trained on distortions, $\mathcal{H}$ is trained only on original CIFAR-10 training data. The results of the proposed contrastive inference compared to feed-forward inference of the augmented model $f'()$ increases by a total of $1.12\%$ across all distortions and levels. On the highest level $5$ distortion in blur category, the increase is $6.87\%$.
\vspace{-3mm}
\subsubsection{Results on STL dataset}
The proposed approach is implemented on higher resolution images of size $96\times96\times3$ in STL-10 dataset~\cite{coates2011analysis}. ResNet-18 architecture is adopted with an extra linear layer to account for change in resolution. Note that STL-10 does not have a standardised distorted version. Hence, we use the same distortions from Section~\ref{subsubsec:NoiseTrain} to corrupt the STL-10 testset. The results of contrastive inference increases by an average of $2.56\%$ in all but underexposure distortion. In underexposure, the accuracy drops by $1.05\%$. In level $5$ of both blur categories, the contrastive performance gain is $6.89\%$. The decrease in performance in underexposure distortion can be attributed to the change in low level statistical characteristics that are discarded by the network's initial layers. 
\vspace{-3mm}
\subsubsection{Results on VisDA dataset}\label{subsubsec:DA}
We show validation results on a synthetic-to-real domain shift dataset called VisDA~\cite{peng2017visda} in Table~\ref{table:VisDA}. An ImageNet pre-trained ResNet-$18$ architecture is finetuned on synthetically generated images from VisDA dataset and tested on real images in the same VisDA dataset. The dataset has $12$ classes with $152k$ training images. While there is an overall performance gain of $5.48\%$, the individual class accuracies indicate room for improvement. The feed-forward predictions, $P$, on \texttt{Knife}, \texttt{Person}, \texttt{Bike}, and \texttt{Train} are either too confidant or their confidence is low. Hence, $Pr(i|P)$ term in Eq.~\ref{eq:probability} is adversely affected by $P$ which in turn affects contrastive predictions $\Tilde{y}$.

\section{Conclusion}
\label{Sec:Discussion}
In this paper, we illustrate the existence of implicit contrast within trained neural networks. We abduce this contrast and provide a robust inference scheme that reasons contrastively. We also visualize such a posteriori reasons as visual explanations that add context to existing causal explanations. The underlying principle that allows extracting such contrastive reasons is : 1) the definition of contrast which allows a datapoint to adopt multiple labels and exist in multiple manifolds and 2) existence of gradients that provide a non-heuristic means to travel between such manifolds.

\bibliographystyle{ieee_fullname}
\bibliography{references}

%
\vspace{-1cm}
\begin{IEEEbiography}[{\includegraphics[width=1in,height=1in,clip,keepaspectratio]{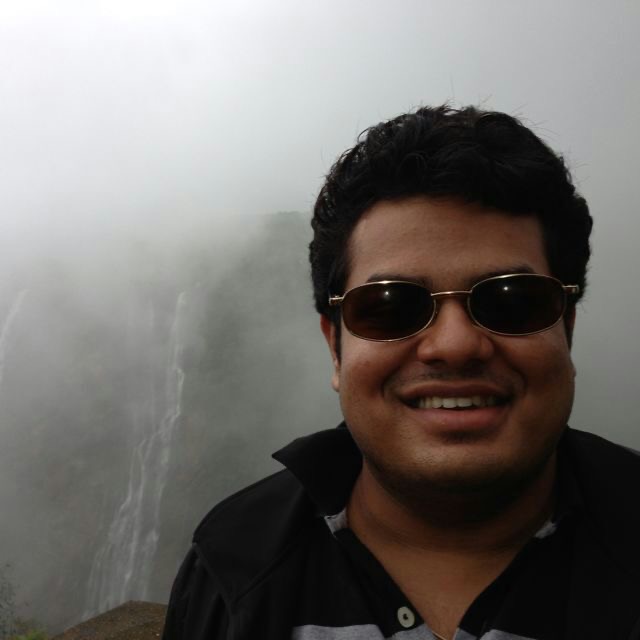}}]{Mohit Prabhushankar}
Mohit Prabhushankar received the B.S Degree in Electronics and Communications Engineering from PES Institute of Technology, Bangalore, India in 2014 and the M.S Degree in Electrical Engineering from Georgia Institute of Technology, Atlanta, GA, USA, in 2015. Since then, he has been a Ph.D student in the Omni Lab for Intelligent Visual Engineering and Science (OLIVES) lab, headed by Ghassan AlRegib, working in the fields of image processing and machine learning. He is the recipient of the Best Paper award at ICIP 2019 and Top Viewed Special Session Paper Award at ICIP 2020. He has served as a Teaching Fellow at Georgia Tech since 2020.
\end{IEEEbiography}
\vspace{-1cm}

\begin{IEEEbiography}[{\includegraphics[width=1in,height=1in,clip,keepaspectratio]{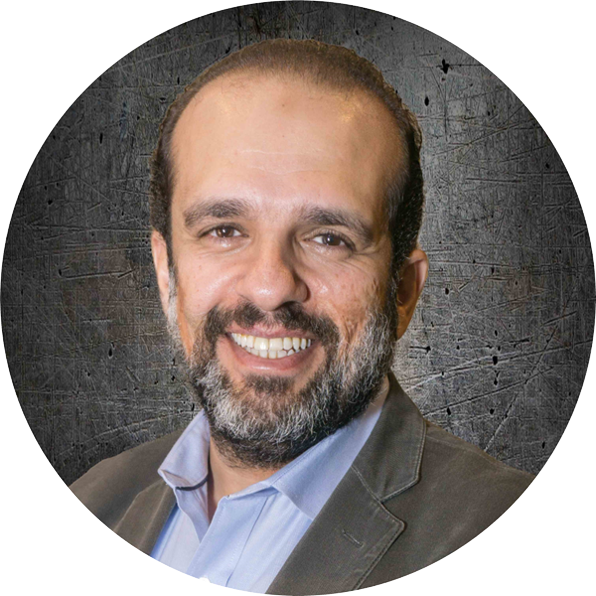}}]{Ghassan AlRegib}
Ghassan AlRegib is currently a Professor with the School of Electrical and Computer Engineering, Georgia Institute of Technology. He was a recipient of the ECE Outstanding Graduate Teaching Award in 2001 and both the CSIP Research and the CSIP Service Awards in 2003, the ECE Outstanding Junior Faculty Member Award, in 2008, and the 2017 Denning Faculty Award for Global Engagement. His research group, the Omni Lab for Intelligent Visual Engineering and Science (OLIVES) works
on research projects related to machine learning, image and video processing, image and video understanding, seismic interpretation, healthcare intelligence, machine learning for ophthalmology, and video analytics. He participated in several service activities within the IEEE including the organization of the First IEEE VIP Cup (2017), Area Editor for the IEEE Signal Processing Magazine, and the Technical Program Chair of GlobalSIP14 and ICIP20. He has provided services and consultation to several firms, companies, and international educational and Research and Development organizations. He has been a witness expert in a number of patents infringement cases.
\end{IEEEbiography}

\end{document}